\pdfoutput=1

\documentclass[11pt]{article}

\usepackage[final]{acl}
\usepackage{booktabs}
\usepackage{times}
\usepackage{amsmath}

\usepackage{latexsym}
\usepackage{graphicx}
\usepackage{authblk}
\usepackage[T1]{fontenc}
\usepackage{tikz}
\def\checkmark{\tikz\fill[scale=0.4](0,.35) -- (.25,0) -- (1,.7) -- (.25,.15) -- cycle;} 
\usepackage{multirow}

\usepackage[utf8]{inputenc}

\usepackage{microtype}

\usepackage{inconsolata}

\title{Leveraging Prompt-Learning for Structured Information Extraction from Crohn’s Disease Radiology Reports in a Low-Resource Language}

\author[1]{\textbf{Liam Hazan}}
\author[5]{\textbf{Gili Focht}}
\author[2]{\textbf{Naama Gavrielov}}
\author[1]{\textbf{Roi Reichart}}
\author[3]{ \textbf{Talar Hagopian}}
\author[4]{\\ \textbf{Mary-Louise C. Greer}}
\author[3]{\textbf{Ruth Cytter Kuint}}
\author[5]{\textbf{Dan Turner}}
\author[2]{\textbf{Moti Freiman}}

\affil[1]{Faculty of Data and Decision Sciences, Technion - Israel Institute of Technology, Haifa, Israel}
\affil[2]{Faculty of Biomedical Engineering, Technion - Israel Institute of Technology, Haifa, Israel}
\affil[3]{Department of Radiology, Shaare Zedek Medical Center, Jerusalem, Israel}
\affil[4]{Department of Radiology, Hospital for Sick Children, Toronto, Canada}
\affil[5]{The Juliet Keidan Institute of Pediatric Gastroenterology, \protect\\ Shaare Zedek Medical Center, Jerusalem, Israel}
\begin{document}
\maketitle
\begin{abstract}

Automatic conversion of free-text radiology reports into structured data using Natural Language Processing (NLP) techniques is crucial for analyzing diseases on a large scale. While effective for tasks in widely spoken languages like English, generative large language models (LLMs) typically underperform with less common languages and can pose potential risks to patient privacy. Fine-tuning local NLP models is hindered by the skewed nature of real-world medical datasets, where rare findings represent a significant data imbalance. We introduce SMP-BERT, a novel prompt learning method that leverages the structured nature of reports to overcome these challenges. In our studies involving a substantial collection of Crohn's disease radiology reports in Hebrew (over 8,000 patients and 10,000 reports), SMP-BERT greatly surpassed traditional fine-tuning methods in performance, notably in detecting infrequent conditions (AUC: 0.99 vs 0.94, F1: 0.84 vs 0.34). SMP-BERT empowers more accurate AI diagnostics available for low-resource languages.
\end{abstract}

\section{Introduction}
Medical imaging, particularly Computed Tomography (CT) and Magnetic Resonance Imaging (MRI), emerges as a key element in the management of complex conditions such as Crohn's Disease (CD) \citep{minordi2022ct} serving as a cornerstone for diagnosis, monitoring, and guiding treatment decisions \citep{Bruiningctmrirecomend}.  Large-scale analyses of imaging data in CD hold promise for advancing research on the inflammatory burden in the bowel and developing predictive models of disease progression \citep{gu2024ai}. The critical clinical information extracted from these images is typically embedded in free-text radiology reports, presenting a significant challenge for large-scale analysis. 

Manually extracting phenotypes and other pertinent information from radiology reports is labor-intensive and requires domain-specific expertise in radiology.  Furthermore, CD exhibits high heterogeneity in the disease course, necessitating manual evaluation of a wide range of potential conditions \citep{Torres2017}.  This task's time-consuming nature and impracticality for large-scale applications pose significant challenges in achieving efficient and accurate data extraction. 

\begin{figure}[t]
    \centering
    \includegraphics[width=0.48\textwidth]{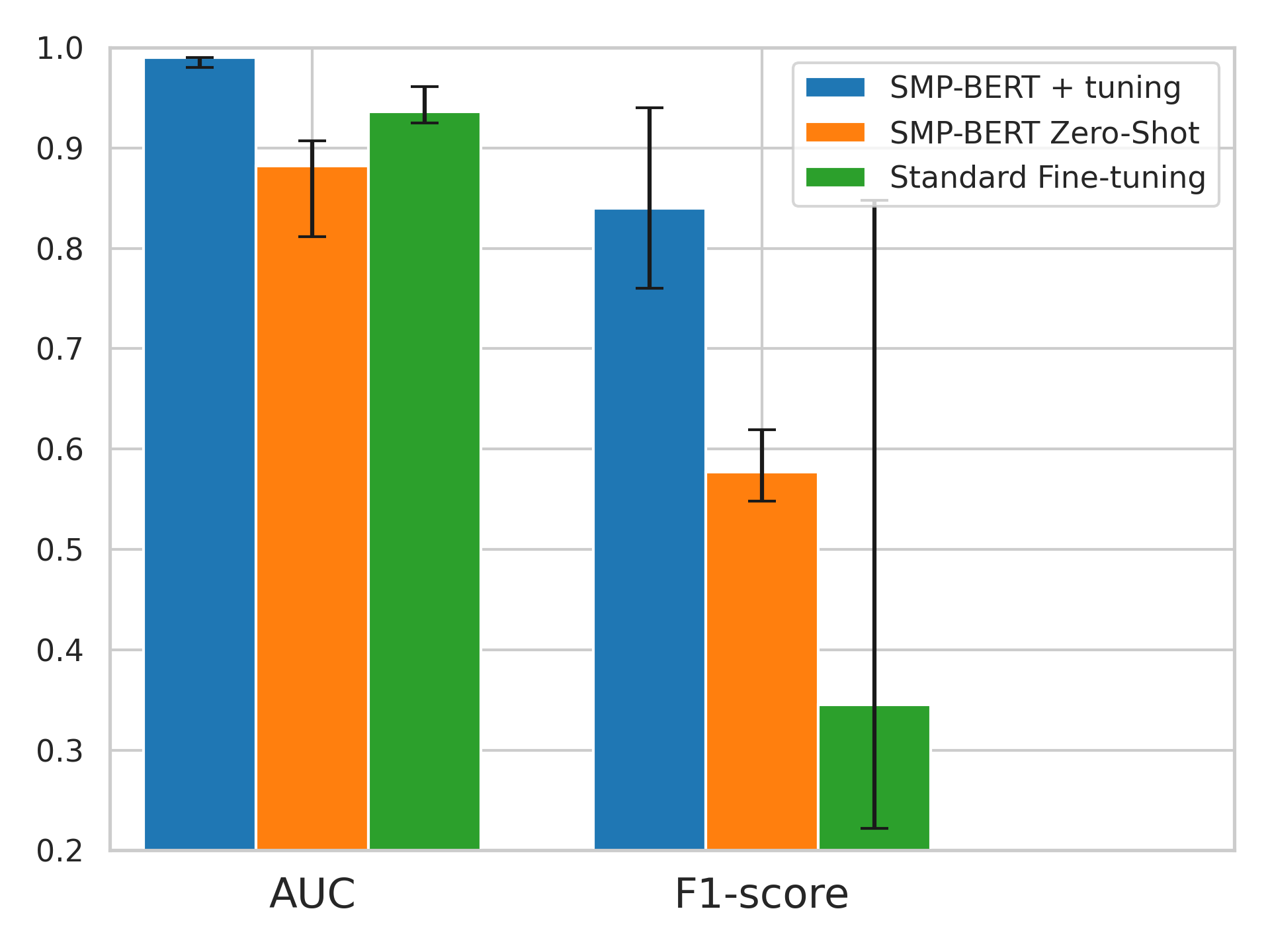}
    \caption{Comparison of the median AUC and F1-score of three models (Standard Fine-tuning, SMP-BERT Zero-Shot, and SMP-BERT + tuning) over all phenotypes with 10+ positives. Error bars represent the Interquartile Range (IQR).}
    \label{fig:model_f1_comparison}
\end{figure}
\begin{figure*}[t]
    \centering
    \includegraphics[width=1\textwidth]{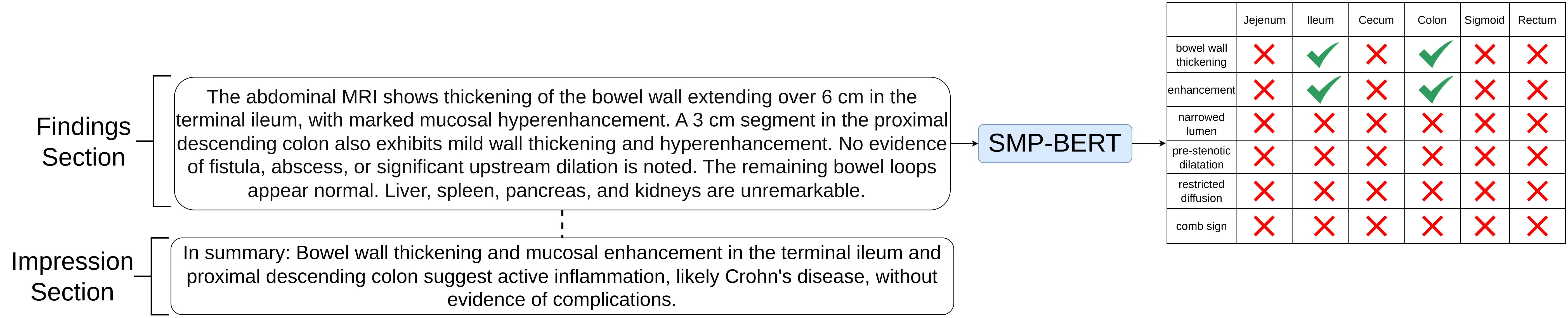}
    \caption{Example of SMP-BERT Input and Output. A medical radiology report section relevant to a patient's CD diagnosis. The section labeled ``Findings'' serves as the input for the SMP-BERT model, similar to its pre-training phase.}
    \label{fig:output_showcase}
\end{figure*}
Recent attempts to automate this extraction process have utilized generative Large Language Models (LLMs) such as GPT-4, which leverage free-text instructions instead of requiring annotated data for training \citep{GPT-4inRadiology}. While these models hold promise, concerns regarding low-resource languages and data privacy remain a challenge.   

Other approaches have involved directly fine-tuning open-source language models on a manually labeled subset of the data \citep{smit2020chexbert, yan2022radbert}.  However, fine-tuning performance suffers from significant data imbalance, a common challenge in medical datasets and particularly in the case of CD, which features some rare conditions. 

To address these limitations, we propose SMP-BERT, a novel prompt learning method built upon the ``pre-train, prompt, and predict'' framework \citep{liu2023pre}, specifically tailored for the structured nature of radiology reports. SMP-BERT leverages a new pre-training task called Section Matching Prediction (SMP). This task leverages the structured format of radiology reports, where key findings reside in some ``Impression'' section. By pre-training on this task, SMP-BERT can infer in a zero-shot setting and also further fine-tune using a relatively small amount of annotated data. This approach not only mitigates the challenge of data imbalance but also eliminates the need for massive training corpora during pre-training.  This advantage makes SMP-BERT readily applicable to low-resource languages, paving the way for a more inclusive and efficient method of extracting information from radiology reports.

\section{Related Work}
\subsection{Radiology Reports Information Extraction}
Various natural language processing approaches have been used in the past to extract information and identify findings on radiology reports, from rule-based methods to deep learning–based language models \citep{smit2020chexbert,mozayan2021practical,tejani2022performance,fink2022deep}. While deep learning models like ClinicalBERT \citep{huang2019clinicalbert}, and RadBERT \citep{yan2022radbert} exploited the use of pre-training on clinical notes and radiology reports, they still require human annotation and a somewhat balanced dataset for fine-tuning. 

Generative LLMs, such as GPT-4 and Cluade, may have clear advantages: They don’t require extra training and can be easily instructed in natural language to do the task with high performance \citep{GPT-4inRadiology}. Unfortunately, radiology reports are usually confidential and can’t be sent as a query through the Internet. Although open-source LLMs might be the solution \citep{mukherjee2023feasibility} they are still focused on English and struggle when it comes to low-resource languages. Moreover, even GPT4 gets comparable results to those of fine-tuned BERT in German \citep{Adams2023} and an open-source model Vicuna-13B also gets comparable results to BERT-based model \citep{mukherjee2023feasibility}.

\subsection{Prompt Learning}

Prompt learning \citep{liu2023pre} is a recent advancement in Natural Language Processing (NLP) that offers a powerful alternative to traditional supervised learning methods which rely on extensive datasets for training a model $P(y|x; \theta)$. Utilizing pre-trained language models (LMs), this approach employs specific input prompts to extend the models' capabilities to tasks beyond their original training. It capitalizes on the input text's probability $P(x; \theta)$, enabling effective use of the comprehensive knowledge amassed by LMs during pre-training. Prompt learning's benefits include its efficient use of data, versatility across different tasks, and reduced need for additional extensive training.

Most prompt learning techniques are based on token-level pre-training tasks such as Left-to-Right Language Modeling \citep{radford2019language,brown2020language} or Masked Language Modeling \citep{PET,schick2020s}. However, a handful of approaches operate at the sentence level, such as \citep{wang2021entailment}, which reformulates the classification task into an entailment task between two sentences.

NSP-BERT \citep{NSP-BERT} is another technique that employs sentence-level pre-training through the Next Sentence Prediction (NSP) task. It uses a structured input format beginning with a \texttt{[CLS]} token, followed by two sentences, A and B, separated by a \texttt{[SEP]} token. The training model balances instances where B genuinely follows A (IsNext) with cases where B is a random sentence (NotNext). The NSP component predicts the likelihood of B following A, relying on a specific matrix $\text{W}_{nsp}$ and the \texttt{[CLS]} token's hidden vector. For tasks like sentiment analysis, one might use a sentence such as ``The ambiance of the restaurant was cozy and inviting,'' and assess if the sentiment is positive by juxtaposing it with prompts like ``The sentiment of this sentence is positive.'' and ``The sentiment of this sentence is negative.'', comparing their ``IsNext'' probabilities. This approach allows labels to correspond with phrases of varying lengths, crucial for extracting information from radiology reports, which often contain findings described in multiple words.

\begin{figure*}[t]
    \centering
    \includegraphics[width=1\textwidth]{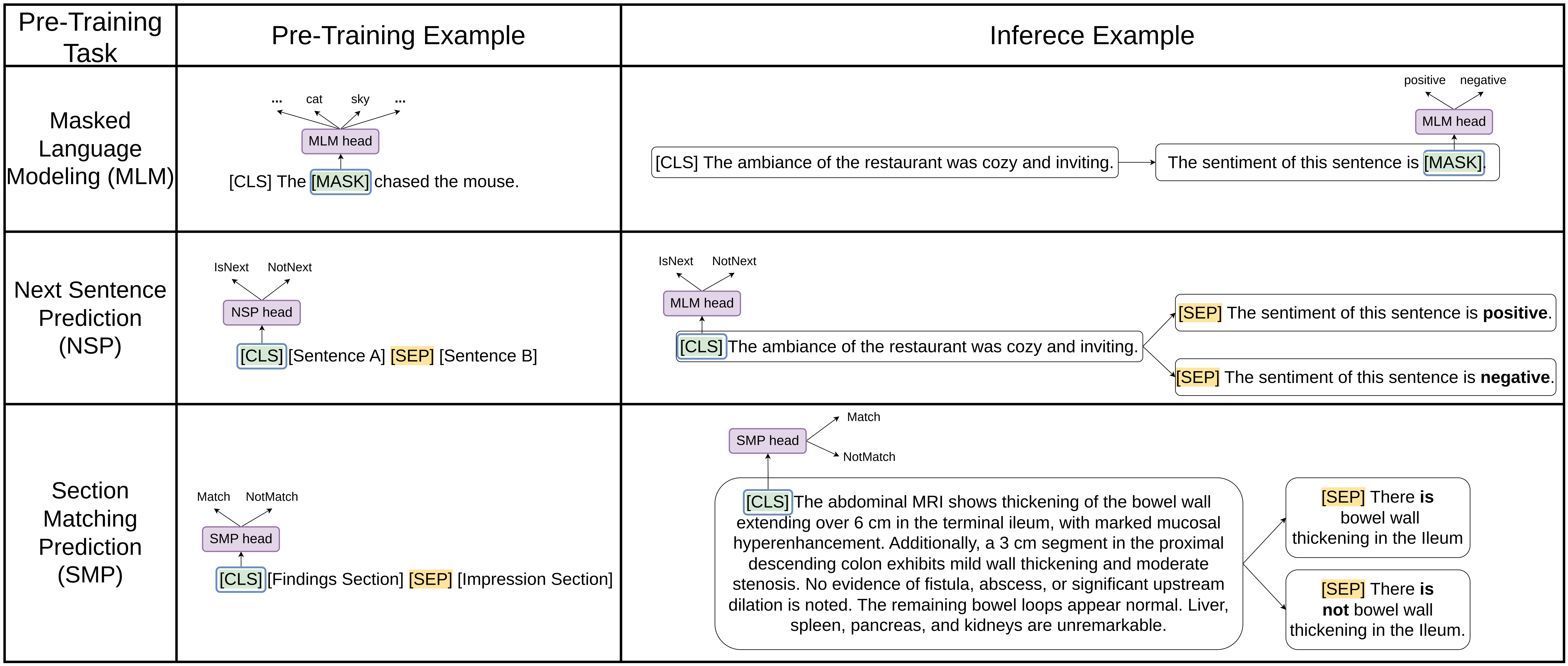}
    \caption{SMP-BERT Methodology - This figure illustrates three pre-training tasks and how they can be used for text classification through prompt learning. Using MLM (token-level) for inference requires ``cloze question'' prompts and a verbalizer function to convert labels into single-token answers (e.g., ``positive''/``negative''). Using NSP (sentence-level) is more simple. While it allows prompts of varying lengths, it's still limited to single-sentence classification. Our novel SMP solves it by pre-training on matching whole sections (multiple sentence level). Then, replace the ``Impression'' section with a prompt about the presence/absence of a finding.  }
    \label{fig:methods}
\end{figure*}

NSP-BERT is optimized for classifying individual sentences, as demonstrated in the pre-training task \ref{fig:methods}. However, radiology reports consist of multiple sentences, posing a challenge for its application. Furthermore, NSP-BERT capitalizes on the logical progression found in narrative texts, where the sequence of ideas or events aids in making predictions. Contrarily, radiology reports primarily present factual details without a narrative flow, diminishing the method's effectiveness in such contexts.

\section{SMP-BERT Framework}
\subsection{Section Matching Prediction}
 
To overcome these challenges, we propose the Section Matching Prediction (SMP) task, designed specifically for analyzing radiology reports. These reports typically contain structured sections, notably ``Findings'' and ``Impression''. The ``Findings'' segment provides detailed observations from radiological examinations, while the ``Impression'' segment offers crucial observations and their summarized interpretations. SMP, inspired by the Next Sentence Prediction approach, considers ``Findings'' as the first segment and ``Impression'' as the follow-up. During training, ``Impression'' sections are accurately matched with their ``Findings'' counterparts half of the time (\texttt{Match}), and mismatched the rest (\texttt{NotMatch}).

Let $\mathcal{M}$ denote the model trained on our radiology reports. The model is trained on the SMP task where $x^F$ and $x^I$ represent the findings and impression sections, respectively. The model's input takes the following form:

$x_{input} = \texttt{[CLS]} x_i^F \texttt{[SEP]} x_i^I \texttt{[EOS]}$

Let $q_\mathcal{M}(n_k|x_i^F,x_i^I)$ denotes the output probability from the model's SMP head based on the input, where $n \in \{\texttt{Match, NotMatch}\}$. The scores $s$ are computed by: $s= \text{W}_{smp} (\text{Tanh}(\text{W}\text{h}_{\texttt{[CLS]}}+\text{b}))$ where $\text{h}_{\texttt{[CLS]}}$ represents the hidden vector of the special token \texttt{[CLS]} and $\text{W}_{smp}$ is the SMP head matrix. The output probability is calculated using the softmax function:
$$q_\mathcal{M}(n_k|x_i^F,x_i^I) = \frac{\text{exp}\: s(n_k|x_i^F,x_i^I)}{\sum_n \text{exp}\: s(n|x_i^F,x_i^I) } $$
This training process, optimized by a cross-entropy loss function, allows the model to discern and assess the logical link between these report sections effectively. During inference, we can leverage this learned ability to construct prompts that specifically target the presence or absence of findings in our reports.

\subsection{Inference with SMP-BERT}
In the inference stage, SMP-BERT leverages its pre-trained understanding of the connection between ``Findings'' and ``Impression'' sections. We substitute the ``Impression'' section with a prompt corresponding to the presence/absence of a clinical finding. By analyzing both the ``Findings'' section and the prompt, SMP-BERT assigns a higher probability to ``\texttt{Match}''" when the prompt aligns with the content of the ``Findings'' section. The input for inference is formulated as: $x_{input} = \texttt{[CLS]} x_i^F \texttt{[SEP]} p^{j} \texttt{[EOS]}$. Here, $p^{j}$ represents the prompt corresponding to the j'th label (presence/absence of a finding).


The template $\mathcal{T}$ combines the report's findings section $(x_i^F)$  with generalized prompt:
$\mathcal{T}(x) =$ \texttt{[CLS]} $x^F$ \texttt{[SEP]} \text{There \{is/isn't\} \{finding\}} \text{in the \{organ\}} \texttt{[EOS]}.
 This approach maps labels to prompts of varying lengths. A verbalizer function $f : \mathcal{Y} \to \mathcal{P}$ associates each label $y^{j} \in \mathcal{Y}$ with its corresponding prompt $p^{j} \in \mathcal{P}$.
 For example, let $p^j = \text{``There is narrowed lumen in the Ileum''}$ and $p^k = \text{``There is \textbf{not} narrowed lumen in the Ileum''}$ then, the prediction for report $x_i$ regarding narrowed lumen in the Ileum would be $\texttt{argmax} \: ({q_\mathcal{M}(\texttt{Match}|x_i^F,p^k), q_\mathcal{M}(\texttt{Match}|x_i^F,p^j)})$.

\subsection{SMP-tuning}
\begin{figure*}[!h]
    \centering
    \includegraphics[width=0.8\textwidth]{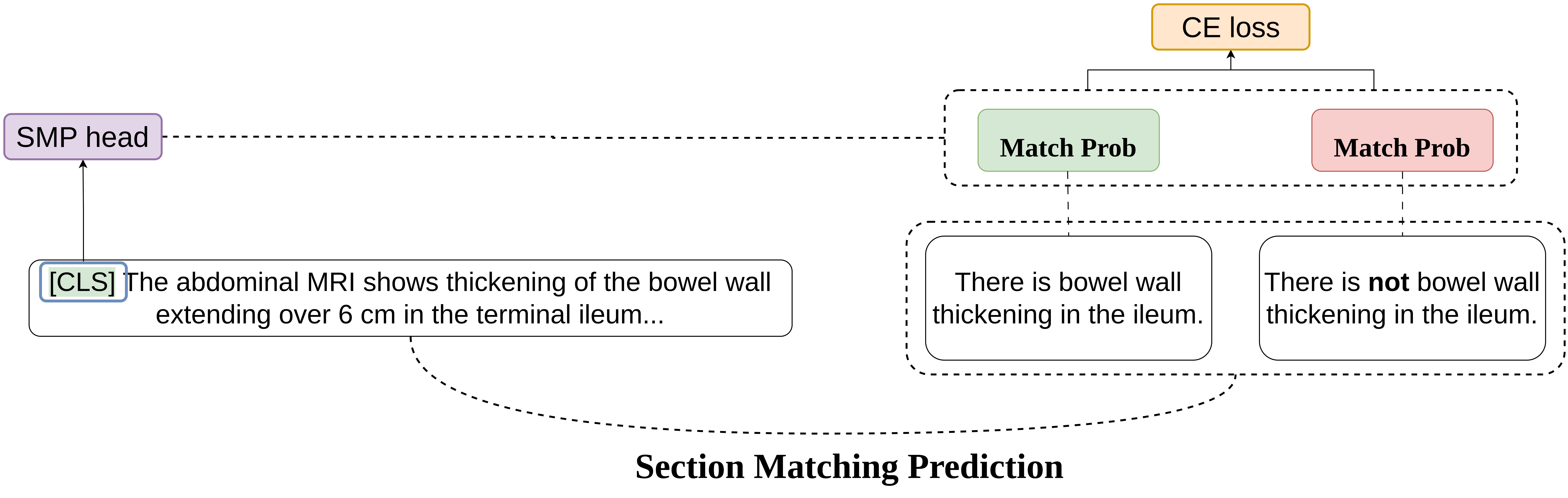}
    \caption{SMP-tuning - Fine-tuning SMP-BERT by generating a negative and a positive instance for every annotated sample and every label. The true label is ``There is {finding} ...'' so the negative instance is paired with ``There is not {finding} ...''}
    \label{fig:smp-tuning}
\end{figure*}

The SMP-tuning process is visualized in Figure \ref{fig:smp-tuning} and conducted similarly to the approach of NSP-tuning from NSP-BERT \citep{NSP-BERT}.

Generally, this process is a continuation of the SMP pre-training just given annotated reports we use the prompts instead of actual ``Impression'' sections. 
Given a sample $i$ with its reference label $y_i^+$, we define a positive instance as $(\mathcal{T}(x_i, y_i^+), \texttt{Match})$ and for each label $y_i^-$ that does not match the reference label, we define negative instances as $\{(T(x_i, y_i^-), \texttt{NotMatch})\}_{y_i^- \in Y \setminus \{y_i^+\}}$, where $Y$ is the set of all possible labels. 
This constructed data sums up to ($\texttt{n\_samples*n\_phenotypes*n\_labels}$) instances and then used to fine-tune the model, leveraging the initialized weights from the SMP pre-training phase.

\section{Experiments}

\begin{figure*}[t]
    \centering
    \includegraphics[width=1\textwidth]{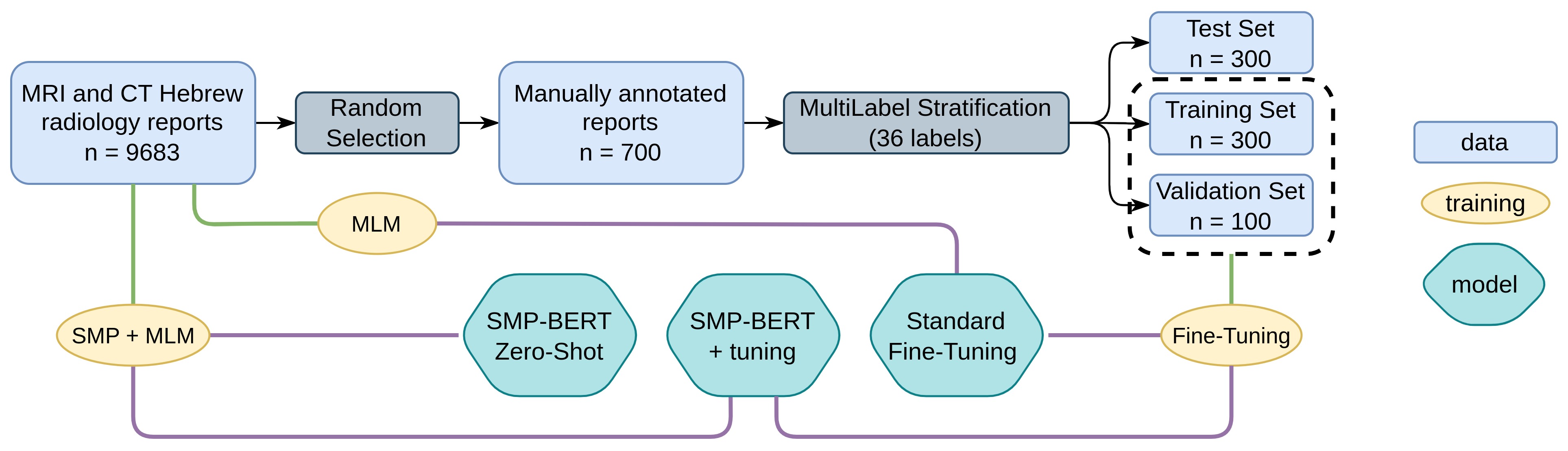}
    \caption{Flowchart of study design - The flowchart outlines the sequence of processing steps from data acquisition to model evaluation. It visualizes the progression from the initial collection of MRI and CT Hebrew radiology reports, through the stages of manual annotation and multi-label stratification, culminating in the pre-training/training of the different models. }
    \label{fig:data+models}
\end{figure*}
\subsection{Data}
This study's dataset consists of radiology reports from three medical institutions, spanning 2010 to 2023. This dataset contains 9,683 free-text reports (one for each visit) for 8093 distinct patients. Since this dataset is confidential, no study has used it to assess the performance of any model.
Ethics approval was obtained from the Shaare Zedek Medical Center Institutional Review Board (Helsinki) committee. 

 For this study, a subset of 700 reports were manually annotated for the presence or absence of certain phenotypes in various organs according to the Consensus Recommendations of the American Gastroenterological Association and the Society for Abdominal Radiology \cite{Bruiningctmrirecomend}. The annotations focused on the following organs: organs jejunum, ileum, cecum, colon, sigmoid, and rectum.  Specific findings annotated included bowel wall thickening, hyper-enhancement, pre-stenotic dilatation, narrowed lumen, restricted diffusion, and comb sign. 
 Since our radiology reports are in the form of free text, we segmented them into ``Findings'' and ``Impression'' sections using keywords like ``In summary:".
 
 \subsection{Experimental Setup}
   We divided the dataset into three distinct sets using a multi-label stratification \citep{sechidis2011stratification}: training (300 reports), validation (100 reports), and test (300 reports) as illustrated in Figure \ref{fig:data+models}. This stratification was crucial to maintain representative distributions of labels across the sets, considering the significant class imbalance present in the majority of labels. 

 Our goal was to compare the performance of our method against standard fine-tuning and assess the advantages of adding the SMP-tuning step on top of the zero-shot approach. 

The foundation of our models is the Hebrew RoBERTa (HeRo) model \citep{shalumov2023hero}, initially pre-trained on the HeDC4 corpus, a comprehensive Hebrew language corpus. We further pre-trained the model on all our radiology reports using the Masked Language Modeling (MLM) task, since there are no other open medical large corpora for Hebrew. 

We conducted experiments using three models: 
\begin{itemize}

    \item \textbf{Standard Fine-tuning}: This model was fine-tuned directly for multi-label classification for all phenotypes.  
    \item \textbf{SMP-BERT Zero-Shot}: This model was further pre-trained on all radiology reports using the SMP task. Inference was executed using the SMP-BERT methodology mentioned in the Inference section. 
    \item \textbf{SMP-BERT + tuning}: Like the zero-shot model, this model underwent pre-training with the SMP task on all radiology reports. Additionally, it was trained further using SMP-tuning to optimize its performance. 
\end{itemize}

 In addition, we assessed the impact of training set size: The models were trained on datasets of varying sizes (50 to 300 reports) to analyze how the amount of training data affects their performance and ability to generalize to unseen data. We further conducted an ablation study to asses the contributions of MLM and SMP pre-training tasks to the model's performance.

Our initial goal was to compare our method with open-source generative LLMs like Llama 2. However, currently available open-source LLMs are not optimized for low-resource languages such as Hebrew, which made the comparison infeasible.

 Due to the inherent class imbalance in the dataset, where most labels have a low number of positive samples, we primarily evaluated the models using the F1-score alongside the AUC metric. The F1-score considers both precision and recall, making it well-suited for imbalanced datasets. Additionally, we reported the Interquartile Range (IQR) along with the scores to provide insight into the variability and distribution of model performance across different labels. 

 All experiments were conducted using a single NVIDIA RTX A6000 GPU, with each experiment taking approximately 1-3 hours. 
 \subsubsection*{Hyper-parameters}
For SMP-BERT + tuning, we train 6 epochs on the constructed dataset ($300*36*2=21600$). For standard Fine Tuning, we trained 120 epochs on the original data (300). For both we set learning rate as 2e-5 with linear decay and the batch size is 24.
\begin{figure*}[!ht]
\centering
\begin{minipage}{0.49\textwidth}
    \centering
    \includegraphics[width=\textwidth]{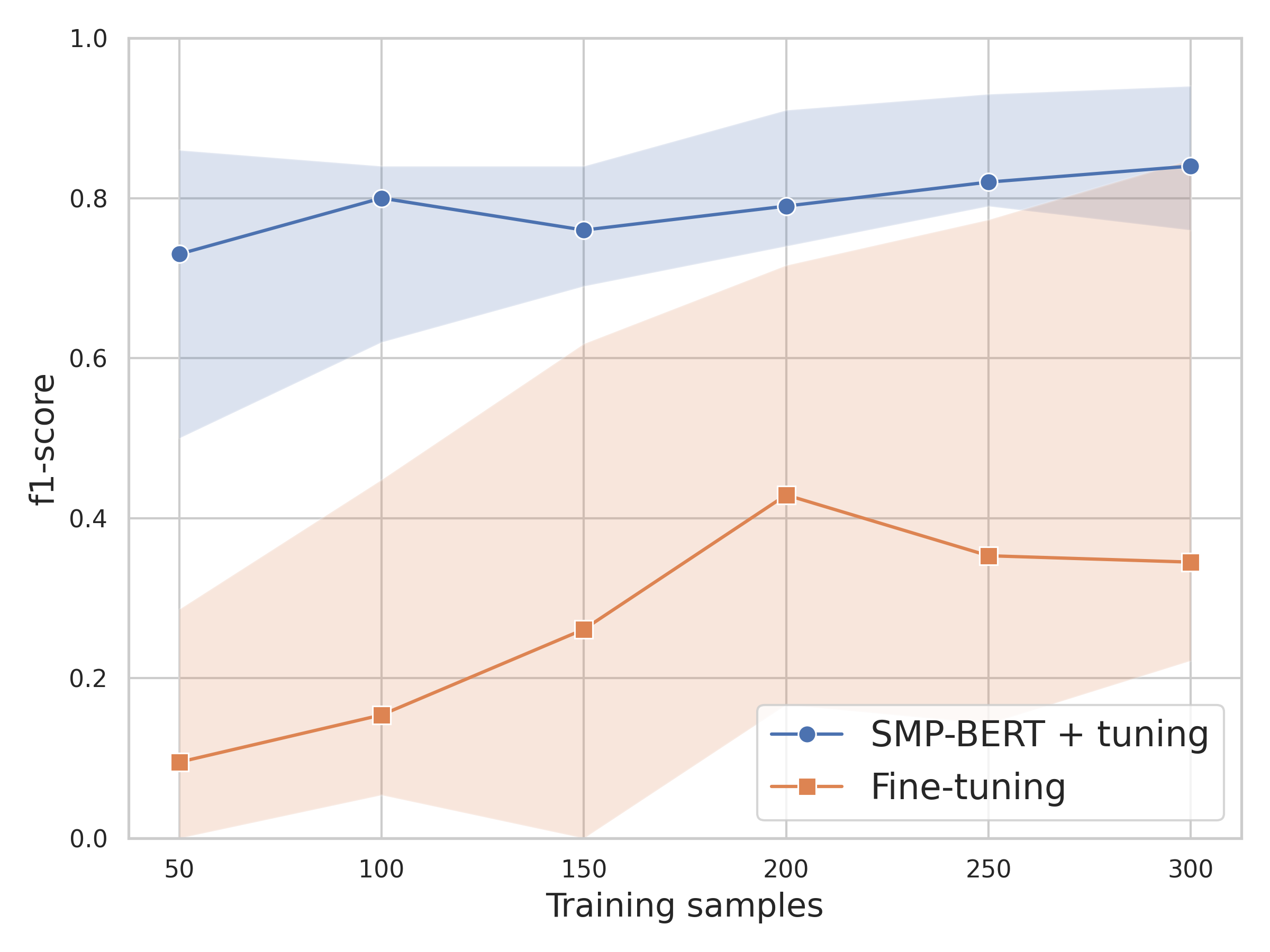}
    \caption{Median F1 scores and IQRs for SMP-BERT + tuning and Standard fine-tuning trained on different training set sizes.}
    \label{fig:training_samples_vs_f1}
\end{minipage}\hfill
\begin{minipage}{0.49\textwidth}
    \centering
    \includegraphics[width=\textwidth]{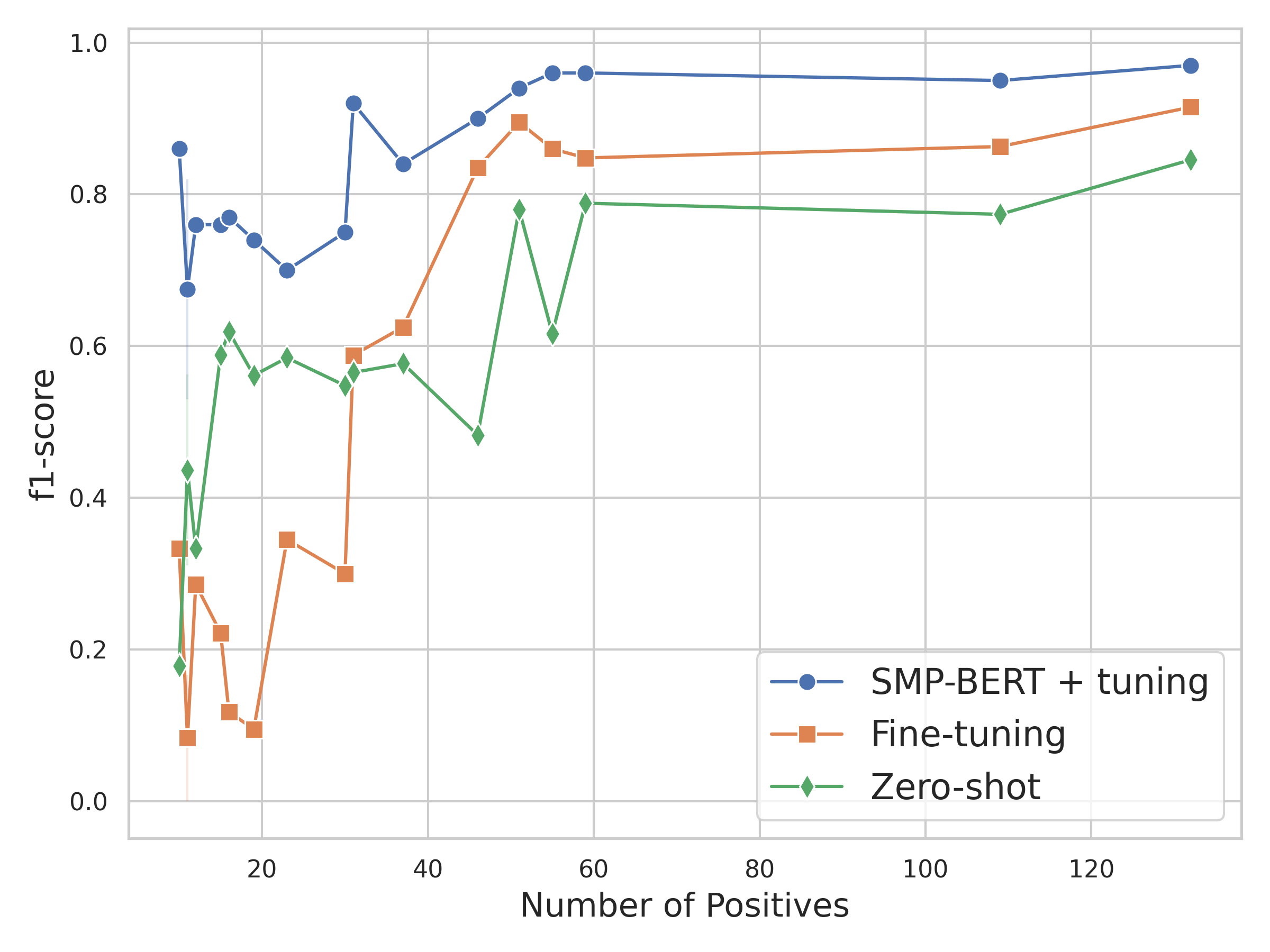}
    \caption{This line chart plots the F1 scores against the number of positive instances of all phenotypes in the dataset (300 total).}
    \label{fig:n_positives_f1}
\end{minipage}
\end{figure*}

\begin{table*}[ht]
    \centering
    \begin{tabular}{|p{4.9cm}|p{2cm}|p{2cm}|p{2.2cm}|p{1.8cm}|}
    \hline
        \textbf{Organ-Finding} & \textbf{SMP-BERT + tuning} & \textbf{SMP-BERT Zero-Shot} & \textbf{Standard Fine-Tuning} & \textbf{prevalence} \\ \hline
        ileum-bowel wall thickening & \textbf{0.97}/\textbf{1.0} & 0.85/0.91 & 0.92/0.98 & 44\% \\ \hline
        ileum-enhancement & \textbf{0.95}/\textbf{0.99} & 0.77/0.86 & 0.86/0.95 & 36\% \\ \hline
        ileum-narrowed lumen & \textbf{0.96}/\textbf{1.0} & 0.79/0.92 & 0.85/0.97 & 19\% \\ \hline
        ileum-dilatation & \textbf{0.96}/\textbf{1.0} & 0.62/0.87 & 0.86/0.96 & 18\% \\ \hline
        ileum-comb sign & \textbf{0.9}/\textbf{0.99} & 0.48/0.81 & 0.84/0.97 & 15\% \\ \hline
        ileum-restricted diffusion & \textbf{0.94}/\textbf{0.99} & 0.78/0.91 & 0.9/\textbf{0.99} & 16\% \\ \hline
        colon-bowel wall thickening & \textbf{0.84}/\textbf{0.98} & 0.58/0.88 & 0.62/0.93 & 12\% \\ \hline
        colon-enhancement & \textbf{0.92}/\textbf{0.99} & 0.57/0.88 & 0.59/0.95 & 9\% \\ \hline
        colon-comb sign & \textbf{0.86}/\textbf{1.0} & 0.18/0.74 & 0.33/0.94 & 3\% \\ \hline
        colon-restricted diffusion & \textbf{0.76}/\textbf{0.98} & 0.33/0.94 & 0.29/0.91 & 3\% \\ \hline
        rectum-bowel wall thickening & \textbf{0.74}/\textbf{0.96} & 0.56/0.89 & 0.1/0.96 & 6\% \\ \hline
        rectum-enhancement & \textbf{0.76}/\textbf{0.98} & 0.59/0.78 & 0.22/0.89 & 5\% \\ \hline
        sigmoid-bowel wall thickening & \textbf{0.75}/\textbf{0.97} & 0.55/0.77 & 0.3/0.9 & 10\% \\ \hline
        sigmoid-enhancement & \textbf{0.7}/\textbf{0.98} & 0.58/0.89 & 0.34/0.88 & 7\% \\ \hline
        sigmoid-comb sign & \textbf{0.53}/\textbf{0.98} & 0.31/0.78 & 0.17/0.93 & 3\% \\ \hline
        cecum-bowel wall thickening & \textbf{0.77}/\textbf{0.98} & 0.62/0.89 & 0.12/0.93 & 5\% \\ \hline
        cecum-enhancement & \textbf{0.82}/\textbf{0.99} & 0.56/0.93 & 0.0/0.92 & 3\% \\ \hline
    \end{tabular}
    \caption{Performance comparison. Values are F1/AUC scores for each model across different phenotypes. The Prevalence column indicates the percentage of test samples in which the phenotype is present.}
    \label{tab:results_table}
\end{table*}

\section{Results}

To account for the inherent class imbalance in our dataset, we focused our analysis on phenotypes with at least 10 positive samples, ensuring the reliability of our findings. 

Our evaluation across three distinct model configurations highlighted the superior performance of the SMP-BERT + tuning approach in extracting phenotypic information from CD radiology reports.  The SMP-BERT + tuning model achieved the highest median AUC of 0.99 (IQR 0.98-0.99), outperforming the Standard Fine-tuning model’s median AUC of 0.94 (IQR 0.92-0.96) and the SMP-BERT Zero-Shot model’s median AUC of 0.88 (IQR 0.81-0.91). For F1-score evaluations, the SMP-BERT + tuning model again leads with a median score of 0.84 (IQR 0.76-0.94), which is substantially higher than the scores of the Standard Fine-tuning model (0.34, IQR 0.22-0.85) and the SMP-BERT Zero-Shot model (0.58, IQR 0.55-0.62). A comprehensive breakdown of these results, including F1 and AUC scores for individual phenotypes, is detailed in the accompanying Table \ref{tab:results_table}. 

Further analysis presented in Figure \ref{fig:n_positives_f1} of model performance relative to the count of positive instances exhibited the strength of SMP-BERT + tuning, particularly for labels with sparse positives in the training set.  For example, with only 19 positive cases for "Rectum Bowel Wall Thickening," SMP-BERT + tuning achieved a significantly higher F1-score (0.74) compared to the standard model (0.1). This demonstrates its superior ability to generalize well from limited data. 

 However, both models performed well when dealing with abundant positive instances. For example, with 137 positives for "Ileum Bowel Wall Thickening" (almost half the dataset), both models achieved good results, with SMP-BERT + tuning maintaining a decent gap (F1-score 0.97 vs. 0.915 for the standard model).  

The graph shown in Figure \ref{fig:n_positives_f1} indicates that the performance gap between the models decreases with an increase in the number of positive instances. This suggests that while SMP-BERT + tuning shines with limited data, it still performs better when more data is available.

We also analyzed how the size of the training set impacts model performance. As shown in Figure \ref{fig:training_samples_vs_f1}, the SMP-BERT + tuning model exhibits superior adaptability. Notably, it achieves good performance even with limited training data (50-100 samples).
The Standard Fine-tuning model exhibits a trend of broadening IQRs and decrease of median score. This could suggests an improving performance for common phenotypes (like Ileum Bowel Wall Thickening) but potentially decreasing performance for rarer ones due to increased data imbalance.
\subsubsection*{Ablation Study}
As evidenced by Table \ref{tab:ablation_table}, both pre-training tasks, MLM and SMP, significantly contribute to optimizing the performance of SMP-BERT. Moreover, it appears that standard fine-tuning benefits from the inclusion of the SMP task.

\begin{table*}[ht]
\centering
\begin{tabular}{lcccc}
Method & MLM & SMP & F1-Score & AUC \\ \hline
\multirow{4}{*}{SMP-BERT + tuning} & \centering \checkmark & \centering $\checkmark$ & \textbf{0.84 [0.76,0.94]} & \textbf{0.99 [0.98,0.99]} \\ & \centering $\checkmark$ & \centering $\times$ & 0.75 [0.59,0.87] & 0.97 [0.96,0.98] \\ & \centering $\times$ & \centering $\checkmark$ & 0.73 [0.67,0.89] & 0.97 [0.95,0.98] \\ & \centering $\times$ & \centering $\times$ & 0.42 [0.26,0.57] & 0.94 [0.92,0.96] \\ \hline
\multirow{4}{*}{Standard Fine-tuning} & \centering \checkmark & \centering $\checkmark$ & \textbf{0.55 [0.35,0.86]} & \textbf{0.96 [0.95,0.98]} \\ &  \centering $\checkmark$ & \centering $\times$ & 0.34 [0.22,0.85] & 0.94 [0.92,0.96]  \\ & \centering $\times$ & \centering $\checkmark$ & 0.15 [0.0,0.72] & 0.85 [0.82,0.91] \\ &  \centering $\times$ & \centering $\times$ & 0.12 [0.0,0.61] & 0.83 [0.78,0.88]
\end{tabular}
\caption{Ablation Study on Pre-training Tasks.}
\label{tab:ablation_table}

\end{table*}

\section{Discussion}
This study examined the efficacy of SMP-BERT, a novel prompt-learning approach, in extracting detailed information from Hebrew radiology reports of CD patients. Our results reveal that SMP-BERT, especially the fine-tuned version (SMP-BERT + tuning), significantly outperforms the standard fine-tuning approach, , achieving an improvement of 49\% in median F1 score and 5\% in median AUC. 

Our study highlights the significant improvement of SMP-BERT + tuning, achieving superior F1-scores and AUCs compared to standard fine-tuning across all analyzed phenotypes. Notably, the model performs well even with a low amount of annotated data. This improvement is particularly notable for rarer phenotypes, demonstrating the model's ability to handle imbalanced datasets, a common challenge in the medical domain. This robustness is crucial for advancing research in CD and other conditions with diverse clinical presentations.

Furthermore, this study contributes to the growing exploration of prompt learning for NLP tasks in healthcare. Unlike traditional fine-tuning approaches, which require substantial labeled data, SMP-BERT leverages pre-training on the ``Section Matching Prediction'' task and further SMP-tuning to achieve exceptional performance even with limited data. This opens exciting possibilities for applying prompt learning in scenarios with limited annotated data, imbalanced data, or low-resource languages, pushing the boundaries of NLP applications in healthcare. 

\section*{Acknowledgements}
The study was supported by The Leona M. and Harry B. Helmsley Charitable trust.
\bibliography{all}




\end{document}